\documentclass[letterpaper]{article} % DO NOT CHANGE THIS
\usepackage{aaai2026}
\usepackage{times}  % DO NOT CHANGE THIS
\usepackage{helvet}  % DO NOT CHANGE THIS
\usepackage{courier}  % DO NOT CHANGE THIS
\usepackage[hyphens]{url}  % DO NOT CHANGE THIS
\usepackage{graphicx} % DO NOT CHANGE THIS
\usepackage{natbib}  % DO NOT CHANGE THIS AND DO NOT ADD ANY OPTIONS TO IT
\usepackage{caption} % DO NOT CHANGE THIS AND DO NOT ADD ANY OPTIONS TO IT
\usepackage{algorithm}
\usepackage{algorithmic}

\usepackage{newfloat}
\usepackage{listings}
\DeclareCaptionStyle{ruled}{labelfont=normalfont,labelsep=colon,strut=off} % DO NOT CHANGE THIS
\floatstyle{ruled}
\newfloat{listing}{tb}{lst}{}
\floatname{listing}{Listing}
\usepackage{amsmath}
\usepackage{amssymb}
\usepackage{bbold}
\usepackage{booktabs}
\usepackage{multirow}
\usepackage{multicol}
\usepackage{makecell}
\usepackage{colortbl}
\usepackage{xcolor}
\usepackage{pifont}
\newcommand{\cmark}{\ding{51}}
\newcommand{\xmark}{\ding{55}}
\definecolor{lightgray}{rgb}{0.9,0.9,0.9}

\title{Anchor-free Cross-view Object Geo-localization with Gaussian Position Encoding and Cross-view Association}

\author {
    Xingtao Ling,
    Chenlin Fu,
    Yingying Zhu\thanks{Corresponding author.}
}
\affiliations {
    College of Computer Science and Software Engineering, Shenzhen University, China\\
    zhuyy@szu.edu.cn
}

\usepackage{bibentry}
\begin{document}
\nocopyright
\maketitle

\begin{abstract}
Most existing cross-view object geo-localization approaches adopt anchor-based paradigm. Although effective, such methods are inherently constrained by predefined anchors. To eliminate this dependency, we first propose an anchor-free formulation for cross-view object geo-localization, termed AFGeo. AFGeo directly predicts the four directional offsets (left, right, top, bottom) to the ground-truth box for each pixel, thereby localizing the object without any predefined anchors. To obtain a more robust spatial prior, AFGeo incorporates Gaussian Position Encoding (GPE) to model the click point in the query image, mitigating the uncertainty of object position that challenges object localization in cross-view scenarios. In addition, AFGeo incorporates a Cross-view Object Association Module (CVOAM) that relates the same object and its surrounding context across viewpoints, enabling reliable localization under large cross-view appearance gaps. By adopting an anchor-free localization paradigm that integrates GPE and CVOAM with minimal parameter overhead, our model is both lightweight and computationally efficient, achieving state-of-the-art performance on benchmark datasets.

% Most existing cross-view object geo-localization approaches adopt an anchor-based paradigm. Although effective, such methods are inherently constrained by predefined anchors. To overcome this limitation, we propose AFGeo, an end-to-end anchor-free cross-view object geo-localization framework. AFGeo directly predicts the four directional offsets (left, right, top, bottom) to the ground-truth box for each pixel, enabling object localization without any predefined anchors. To further improve localization accuracy, AFGeo integrates Gaussian Position Encoding (GPE) to model spatial priors and a Cross-view Object Association Module (CVOAM) to associate consistent semantics of the object and its surrounding context across views, both introducing minimal additional parameters. This design keeps AFGeo lightweight and computationally efficient, while achieving state-of-the-art performance on benchmark datasets.
\end{abstract}


\section{Introduction}
\label{sec:intro}

\begin{figure}[t]
\centering
\includegraphics[width=\linewidth]{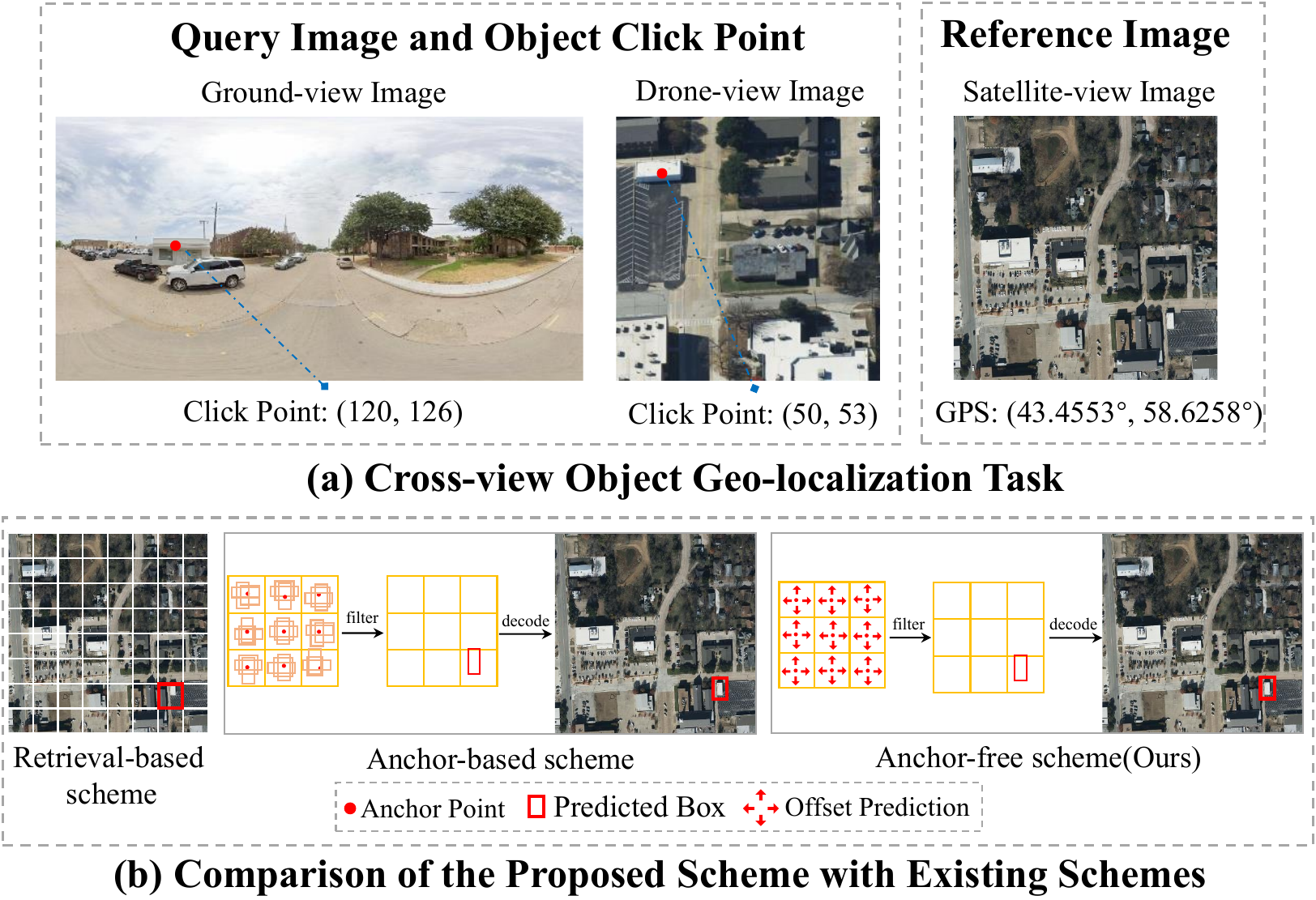}
\caption{Illustration of the cross-view object geo-localization task and comparison of different solutions. \textbf{(a)} The reference image is usually a satellite image with geographic information (a GPS coordinate). The query image is from a ground or drone view, with the click point (red dots) of the object of interest. Cross-view object geo-localization aims to find the geographic location of an object of interest in a query image from a reference image with attached geographic information. \textbf{(b)} The retrieval-based scheme divides the satellite image into uniformly-sized patches to construct a reference database and attempts to retrieve the patch that is most semantically similar to the object of interest as a location proposal. The anchor-based scheme generates a series of candidate bounding boxes for each anchor based on predefined anchors, and then selects the best bounding box containing the object of interest from these candidates. The anchor-free (ours) scheme directly predicts the object location at each pixel and selects the best bounding box containing the object of interest. Obviously, our scheme is no longer constrained by predefined anchors.}
\label{fig:task}
\end{figure}

Consider Fig.~\ref{fig:task} \textbf{(a)}: if we know the location of an object of interest in the query image, how can we find that object in the reference image with known geographic information to obtain the object’s geographic coordinates? Addressing this question is the goal of cross-view object geo-localization, which has important applications in smart city management~\citep{al2015applications, hong2021multimodal}, disaster monitoring~\citep{hansch2022spacenet, zhang2024good}, autonomous driving~\cite{hane20173d} and so on. A straightforward scheme is to imitate cross-view image geo-localization~\citep{workman2015wide, hu2018cvm, shi2019spatial, yang2021cross, zhu2022transgeo, deuser2023sample4geo} using a retrieval-based approach. In this setting, first partition the reference image into uniformly-sized patches and then retrieve the patch most similar to the object of interest as the localization result. However, since objects exhibit diverse shapes and scales, uniformly-sized patches are insufficient to represent objects of arbitrary sizes. Subsequent works~\citep{DBLP:journals/tgrs/SunYKFFLLZP23, 10888758} adopt anchor-based scheme. They adjust a set of predefined anchors with different aspect ratios and select the box that best matches the object as the localization result. This scheme addresses the uniformly-patch limitation.

However, anchor-based paradigm relies on anchors that are manually designed or obtained by clustering. Such anchors neither accurately describe object sizes nor avoid introducing complex hyperparameter tuning and bulky detection heads~\citep{tian2019fcos,kong2020foveabox}. To the best of our knowledge, we are the first to propose an anchor-free localization scheme to overcome these issues in cross-view object geo-localization. In our anchor-free scheme, we draw inspiration from the object detection methodology of FCOS~\citep{tian2019fcos} and directly predict the offsets from each pixel to the object in four directions (left, right, top, bottom), thereby eliminating reliance on prior boxes. Our approach inherits the advantages of anchor-free architectures, including: (1) elimination of dependence on candidate anchors~\citep{tian2019fcos}, leading to reduced computational cost; (2) reduction in the need for hyperparameter tuning~\citep{tian2019fcos,kong2020foveabox}, improving generalization capability; (3) capability to directly predict object locations, enabling more flexible adaptation to objects of different scales. Different cross-view object geo-localization schemes are illustrated in Fig.~\ref{fig:task} \textbf{(b)}.

% Despite the merits of anchor-free scheme, cross-view object geo-localization remains challenging. Query images are typically captured from a near-ground viewpoint and cover a small region containing the object of interest, revealing fine object details. Reference images are typically captured from an aerial viewpoint and cover a large region that includes the object, revealing more of the object’s global appearance. This results in substantial visual discrepancies for the same object across views. Furthermore, the spatial uncertainty of object locations across views makes localization even more difficult.

Despite the merits of anchor-free schemes, cross-view object geo-localization remains challenging. A click point in the query image generally corresponds to an uncertain and spatially dispersed region in the reference image due to projection geometry or annotation imprecision, which makes the task particularly difficult. In addition, owing to the substantial viewpoint differences between query and reference images, as well as variations in capture conditions such as time, illumination, blur, and occlusion, the same object can exhibit significant geometric and appearance discrepancies across views, further complicating cross-view object localization.

To address these challenges, we propose a novel anchor-free cross-view object geo-localization framework. The framework employs a new Gaussian Position Encoding (GPE) that represents the click point in the query image as the center of a 2D Gaussian distribution and uses a learnable standard deviation to adaptively adjust the encoding extent, thereby better accommodating positional uncertainty. In addition, the framework integrates a Cross-view Object Association Module (CVOAM), which associates consistent semantics of the same object and its surrounding context across views via interaction between the module’s two network branches, thereby improving cross-view localization capability. Notably, the two main components of our framework, GPE and CVOAM, introduce no additional parameter overhead (except for the learnable standard deviation in GPE), which contributes to the lightweight nature of the model. The main contributions of this paper are as follows:

% · We first adopt an anchor-free localization paradigm for cross-view object geo-localization. We propose an end-to-end anchor-free cross-view object geo-localization network (AFGeo) that eliminates reliance on candidate anchors and realizes a more lightweight and efficient solution for cross-view object geo-localization.

% · To provide a more robust positional prior, we introduce a novel Gaussian Position Encoding (GPE) that models object location uncertainty as a smooth probability distribution. We also propose a Cross-view Object Association Module (CVOAM) to associate consistent object semantics across views and improve localization performance. Notably, GPE and CVOAM introduce negligible—or even no—additional parameters.

% · Extensive experiments on the CVOGL~\cite{DBLP:journals/tgrs/SunYKFFLLZP23} benchmark and G2D\footnote{https://figshare.com/s/b4002221f7f6eb067fa1} demonstrate that our AFGeo surpasses current state-of-the-art methods, significantly improving localization accuracy while maintaining a compact model size. Our method offers an effective solution for deploying object-level geo-localization in real-world systems with limited computational resources.

\begin{itemize}
    \item We first adopt an anchor-free localization paradigm for cross-view object geo-localization. We propose an end-to-end anchor-free cross-view object geo-localization network (AFGeo) that eliminates reliance on candidate anchors, improving flexibility and generalization capability for cross-view object matching.
    % and realizes a more lightweight and efficient solution for cross-view object geo-localization.
    \item To provide a more robust positional prior, we introduce a novel Gaussian Position Encoding (GPE) that models object location uncertainty as a smooth probability distribution. We also propose a Cross-view Object Association Module (CVOAM) to associate consistent object semantics across views and improve localization performance. Notably, GPE and CVOAM introduce negligible—or even no—additional parameters and contribute to the lightweight design.
    \item Extensive experiments on the CVOGL~\citep{DBLP:journals/tgrs/SunYKFFLLZP23} benchmark and G2D demonstrate that our AFGeo surpasses current state-of-the-art methods, significantly improving localization accuracy while maintaining a compact model size. Our method offers an effective solution for deploying object-level geo-localization in real-world systems with limited computational resources.
\end{itemize}

\section{Related Work}

\textbf{Retrieval-based Cross-view Object Geo-localization.}  
A natural idea for cross-view object geo-localization is to adapt retrieval-based strategies from cross-view image geo-localization~\citep{hu2018cvm, lin2022joint, yang2021cross, shi2019spatial, zhu2022transgeo, deuser2023sample4geo, 10.1007/978-3-031-72630-9_13}. In this setting, the satellite reference image is divided into a set of uniformly sized patches, and each patch is treated as a candidate location. The model extracts feature embeddings for both the query object and all reference patches, and the patch with the highest similarity score is selected as the predicted object location. Although conceptually simple, patch retrieval suffers from several drawbacks. The spatial resolution of localization is inherently limited by the patch granularity, making it difficult to achieve precise bounding box alignment. The exhaustive comparison across all patches also introduces high computational cost for large-scale reference images. These limitations have motivated research into alternative solutions.

\textbf{Anchor-based Cross-view Object Geo-localization.} 
Inspired by advances in object detection, several works~\citep{DBLP:journals/tgrs/SunYKFFLLZP23, 10888758} adopt anchor-based mechanisms for cross-view object localization. These methods generate a large set of predefined anchor boxes in the reference image and then predict classification and regression offsets to identify the target object. While anchor-based strategies achieve better localization accuracy than retrieval-based methods, they suffer from several intrinsic drawbacks. First, detection performance is highly sensitive to the predefined anchor hyperparameters (e.g., sizes, aspect ratios, and numbers), which require extensive heuristic tuning. Second, the fixed design of anchor boxes limits their ability to generalize across objects of varying scales and aspect ratios, particularly in challenging cross-view scenarios. Moreover, dense placement of anchors introduces a massive number of negative samples, leading to severe class imbalance and increased computational overhead. Finally, anchor-based approaches involve additional complexity from IoU computations and anchor-to-ground-truth matching during training.

\textbf{Anchor-free Cross-view Object Geo-localization.}  
Recently, anchor-free approaches~\citep{tian2019fcos, law2018cornernet, duan2019centernet, zhou2019bottom, kong2020foveabox} in object detection have emerged as a promising alternative by eliminating the dependency on predefined anchors. These methods completely sidestep the need for heuristic tuning of hyperparameters related to anchor shapes, sizes, and ratios, which are often critically sensitive to detection accuracy. In this paper, we first adopt an anchor-free paradigm for cross-view object geo-localization. This innovative methodology inherits the intrinsic advantages of anchor-free paradigms, exploring novel possibilities and opening up new avenues for the task of cross-view object geo-localization.

\textbf{Position Encoding \&\& Cross-view Fusion.}
\citet{DBLP:journals/tgrs/SunYKFFLLZP23} propose a positional encoding method based on the Euclidean distance from each spatial location to the click point, generating a spatial weight matrix that highlights the approximate area around the query object This positional prior is concatenated with the query image features and processed through a cross-view fusion module equipped with spatial attention, which enhances the feature discriminability in the reference image by focusing on regions most relevant to the query.~\citet{10888758} introduce a view-specific positional encoding (VSPE) strategy that differentiates between ground and drone views. They further improved feature representation through a channel-spatial hybrid attention (CSHA) mechanism. Unlike these methods, our proposed Gaussian Position Encoding (GPE) represents a significant advancement in handling positional uncertainty and adaptive context modeling. Moreover, we develop the Cross-view Object Association Module (CVOAM), which goes beyond conventional fusion mechanisms by explicitly associating consistent object semantics across views.

\section{Methodology}
\label{sec:method}

\begin{figure*}[htb]
\centering
\includegraphics[width=\linewidth]{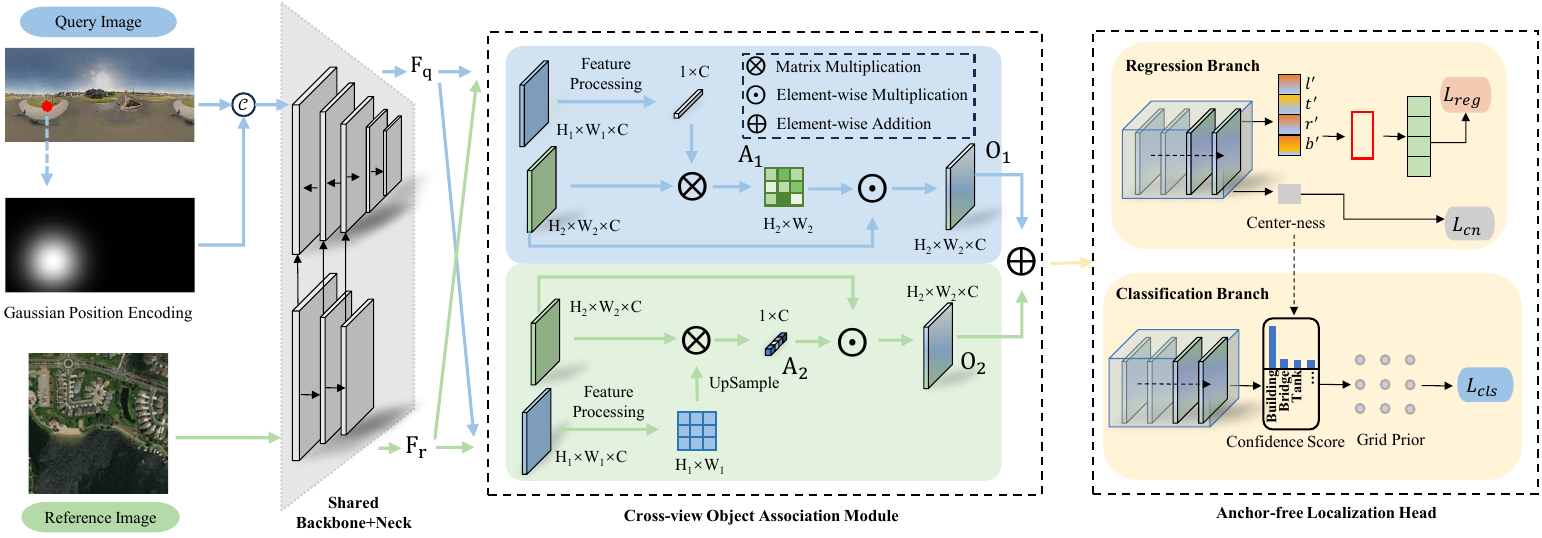}
\caption{Overview of our AFGeo. AFGeo adopts an anchor-free architecture for cross-view object geo-localization. The structure consists of two key components: Gaussian Position Encoding and the Cross-View Object Association Module, with the localization head following the anchor-free paradigm.}
\label{fig:model-structure}
\end{figure*}

The overview of our proposed AFGeo model is shown in Fig.~\ref{fig:model-structure}. The AFGeo architecture mainly consists of Gaussian Position Encoding (GPE), the Cross-View Object Association Module (CVOAM), and an object localization head that adopts the anchor-free paradigm. We will elaborate on the key components in the following sections.

\subsection{Gaussian Position Encoding}
\label{subsec:GPE}

\begin{figure}[htb]
\centering
\includegraphics[width=\linewidth]{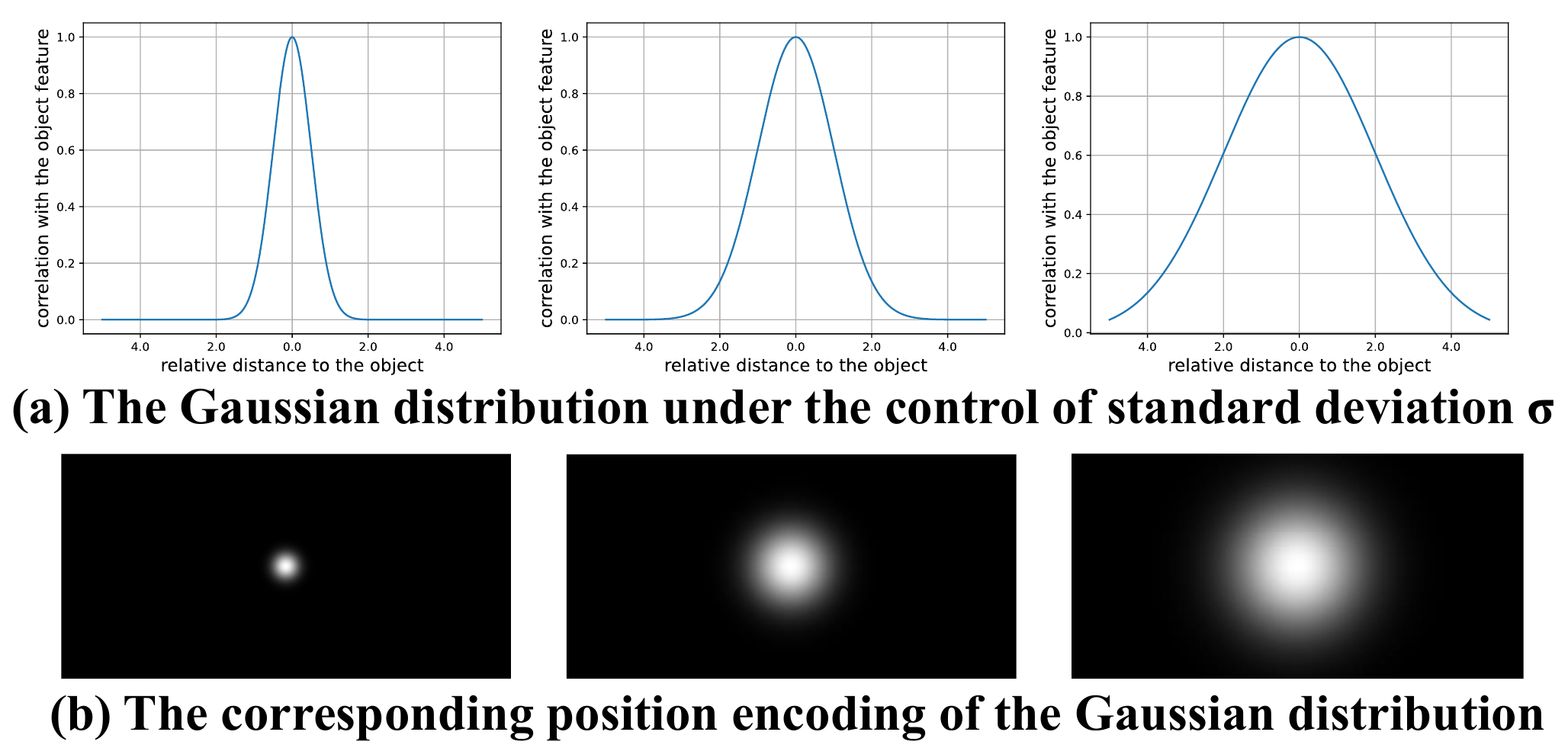}
\caption{Illustration of the proposed GPE. \textbf{(a)} We employ a learnable standard deviation to control the Gaussian distribution, enabling the model to adaptively learn the relationship between the relative distance of the object and the correlation of object features, meaning that the farther the relative distance, the weaker the feature correlation. \textbf{(b)} We visualize the position encodings under different Gaussian distributions. The learnable GPE allows the model to better adapt to the uncertainty of object locations, focusing more on the object feature regions (gray–white areas) while ignoring the interference from irrelevant background (black areas).}
\label{fig:GPE}
\end{figure}

In cross-view object geo-localization, position encoding serves as an effective spatial prior to guide feature extraction and object localization. We observe that the click point and its vicinity contain rich object semantic information. Therefore, reasonable modeling of the click point is crucial for the subsequent task of locating the object in the reference image. In this paper, we model the click point by using Gaussian Position Encoding (GPE), as shown in Eq.~\ref{eq:gaussian}.
\begin{equation}
    \label{eq:gaussian}
    \mathbf{P}_k(i,j) = \exp\left({-\frac{(||z_k(i,j)-p_k||_2)^2}{2\sigma^2}}\right)
\end{equation}
where $\mathbf{P}_k(i,j)$ denotes the encoding of the click point of the $k$-th sample at the $i$-th row and $j$-th column, $z_k(i,j)$ denotes the pixel position at the $i$-th row and $j$-th column of the $k$-th sample, $p_k$ denotes the click point of the object of interest, $\sigma$ denotes the standard deviation. Compared to the Euclidean or hybrid Euclidean distance position encodings in~\cite{DBLP:journals/tgrs/SunYKFFLLZP23, 10888758}, we introduce a learnable standard deviation $\sigma$ in GPE to adaptively control the encoding range. This enables the model to produce a concentrated distribution for small objects and a broader distribution for large objects (as shown in Fig.~\ref{fig:GPE}), providing a more robust spatial prior. Notably, GPE adds only a single $\sigma$ parameter, imposing negligible overhead on the overall model.

\subsection{Cross-view Object Association Module}
\label{subsec:CVOAM}

Human perception offers a crucial insight: when establishing cross-view object associations, observers naturally integrate target objects with their semantic context (e.g., road structure, surrounding buildings). This cognitive behavior indicates that reliable object matching inherently depends on the semantic consistency of the local environment—a critical signal overlooked by existing methods. Current approaches~\cite{deuser2023sample4geo, DBLP:journals/tgrs/SunYKFFLLZP23, 10888758} compress image feature maps into a single global query vector, which overemphasizes holistic representations and results in the loss of fine-grained details. To address this limitation, we introduce a Cross-View Object Association Module (CVOAM) that enables the model to automatically focus on both critical local features and global contextual cues for robust cross-view object matching.

CVOAM consists of two parallel sub-networks that fuse the feature outputs of the query branch $\mathbf{F}_\text{q} \in \mathbb{R}^{H_1 \times W_1 \times C}$ and the reference branch $\mathbf{F}_\text{r} \in \mathbb{R}^{H_2 \times W_2 \times C}$. In the first sub-network (blue background in Fig.~\ref{fig:model-structure}), $\mathbf{F}_\text{q}$ is globally average-pooled over the spatial dimensions and multiplied with raw $\mathbf{F}_\text{r}$ to produce a weight matrix $\mathbf{A}_1 \in \mathbb{R}^{H_2 \times W_2}$. $\mathbf{A}_1$ is then applied to raw $\mathbf{F}_\text{r}$, yielding the sub-network output $\mathbf{O}_1 \in \mathbb{R}^{H_2 \times W_2 \times C}$.
\begin{gather}
    \mathbf{F}_\text{q} = \frac{1}{H_1 \times W_1}\sum^{H_1}_{i=1}\sum^{W_1}_{j=1}\mathbf{F}_\text{q}\left[ i,j,:\right] \\
    \mathbf{A}_1 =  \text{Normalize}\left( \mathbf{F}_\text{r} \cdot \mathbf{F}_\text{q}^{T} \right)\\
    \mathbf{O}_{1} = \mathbf{A}_1 \odot  \mathbf{F}_\text{r}
\end{gather}
Where $\odot$ denotes elment-wise multiplication. Then, in the second sub-network (green background in Fig.~\ref{fig:model-structure}), $\mathbf{F}_\text{q}$ is globally average-pooled over the channel dimensions, upsampled, reshaped and multiplied with reshaped $\mathbf{F}_\text{r}$ to produce a weight matrix $\mathbf{A}_2 \in \mathbb{R}^{1\times C}$. $\mathbf{A}_2$ is then applied to raw $\mathbf{F}_\text{r}$, yielding the sub-network output $\mathbf{O}_2 \in \mathbb{R}^{H_2 \times W_2 \times C}$.
\begin{gather}
    \mathbf{F}_\text{q} = \text{Reshape}( \text{Upsample}(\frac{1}{C}\sum^{C}_{i=1}\mathbf{F}_\text{q}\left[:,:,i\right] ) )\\
    \mathbf{A}_2 =  \text{Reshape}\left( \text{Normalize}\left( \mathbf{F}_\text{q} \cdot \text{Reshape}\left(\mathbf{F}_\text{r}\right) \right) \right)\\
    \mathbf{O}_{2} = \mathbf{A}_2 \odot  \mathbf{F}_\text{r}
\end{gather}
Also, $\odot$ denotes elment-wise multiplication. Finally, the outputs of the two sub-networks are aggregated by element-wise addition to yield the fused representation $\mathbf{O}_\text{fusion} = \mathbf{O}_{1} + \mathbf{O}_{2}$. CVOAM introduces no additional parameters to the model, maintaining the efficiency and compactness of the overall model.

\subsection{Anchor-free Localization Head}
\label{subsec:head}

% 我们在头部分别进行分类预测和回归预测。在分类分支中，模型会预测一个$H \times W \times C_1$的张量，$C_1$ means that we use a 1-D embedding to predict whether a bbox is the target object or not. 在回归分支中，模型会分别预测一个$H \times W \times C_2$和$H \times W \times C_3$的张量。其中，$C_2=1$ means that we use a 1-D embedding to predict中心度信息，用来评估当前像素点作为目标中心的可信度，抑制那些远离真实目标中心的低质量预测；$C_3=4$表示每个像素点到真实框的四个边界距离$(l', t', r', b')$的预测值。

We decouple classification and regression in the anchor-free localization head. The classification branch outputs a tensor $\mathbf{\hat{C}} \in \mathbb{R}^{H \times W \times 1}$, where $\hat{c}_{x,y}$ denotes the targetness logit at location $(x,y)$. Here, the channel dimension is set to 1, as our task focuses on localizing a common foreground object across cross-view images. The regression branch outputs $\mathbf{\hat{S}} \in \mathbb{R}^{H \times W \times 1}$ and $\mathbf{\hat{B}} \in \mathbb{R}^{H \times W \times 4}$. $\hat{s}_{x,y}$ indicates the estimated centerness at location $(x,y)$, which helps suppress low-quality predictions far from the actual object center. $\hat{b}_{x,y}=(\hat{l},\hat{t},\hat{r},\hat{b})$ denotes the predicted distances from location $(x,y)$ to four boundaries of the ground-truth (GT) box. 

\textbf{Training.} Our label assignment strategy follows~\cite{tian2019fcos}. A feature location $(x,y)$ is assigned as positive if it satisfies two criteria: (1) it lies inside a GT box and falls within a disk of radius $\rho$ centered at the GT box center, and (2) the predicted object size falls into the scale range handled by the current feature level. If a location  falls into multiple GT boxes, we simply assign it to the smallest-area box. For a positive location, the targets are $c_{x,y}^{*}=1, b_{x,y}^{*}=(l^*,t^*,r^*,b^*) $ and $s_{x,y}^{*}=\sqrt{\frac{\min \left(l^{*}, r^{*}\right)}{\max \left(l^{*}, r^{*}\right)} \times \frac{\min \left(t^{*}, b^{*}\right)}{\max \left(t^{*}, b^{*}\right)}}$, where $l^*,t^*,r^*$ and $b^*$ are the ground-truth distances from location $(x,y)$ to the ground-truth (GT) box. Locations that do not satisfy the conditions are labeled negative with $c_{x,y}^{*}=0$.

\textbf{Inference.} During inference, the final detection confidence at each location is obtained by multiplying the classification score and the centerness score. The bounding box prediction corresponding to the location with the highest combined confidence is selected as the final output.

\subsection{Loss Function}
\label{subsec:loss}

Our training loss function is defined as follows:
\begin{equation}
\begin{aligned}
    \mathcal{L} = &\frac{\lambda_{cls}}{N_{pos}} \sum_{x,y} \mathcal{L}_{cls} (\hat{c}_{x,y}, c_{x,y}^{*}) \\
    + & \frac{\lambda_{cn}}{N_{pos}} \sum_{x,y} \mathbb{1}_{\left\{c_{x, y}^{*}>0\right\}} \mathcal{L}_{cn} (\hat{s}_{x,y}, s_{x,y}^{*}) \\
    + & \frac{\lambda_{reg}}{N_{pos}} \sum_{x,y} \mathbb{1}_{\left\{c_{x, y}^{*}>0\right\}} \mathcal{L}_{reg} (\hat{b}_{x,y}, b_{x,y}^{*})
\end{aligned}
\end{equation}
Where $\mathcal{L}_{cls}$ is focal loss as in~\cite{lin2017focal}, $\mathcal{L}_{cn}$ is Binary Cross-Entropy (BCE) loss as in~\cite{goodfellow2016deep} and $\mathcal{L}_{reg}$ is GIoU loss as in~\cite{rezatofighi2019generalized}. $\mathbb{1}_{\left\{c_{x, y}^{*}>0\right\}}$ is the indicator function, being 1 if $c_{x, y}^{*} > 0$ and $0$ otherwise. $\lambda_{cls}$, $\lambda_{cn}$ and $\lambda_{reg}$ are balance weights and are set to 1 in this paper. $N_{pos}$ denotes the number of positive samples.

\begin{table*}[t]
\centering
\begin{tabular}{l|cc|cc|cc|cc}
    \toprule
    \multirow{4}{*}{ Method } & \multicolumn{4}{c}{Ground→Satellite} & \multicolumn{4}{c}{Drone→Satellite} \\
    % \cline{2-9}
    \cmidrule(lr){2-5} \cmidrule(lr){6-9}
    & \multicolumn{2}{c}{ Test } & \multicolumn{2}{c}{ Validation } &  \multicolumn{2}{c}{ Test } & \multicolumn{2}{c}{ Validation } \\
    % \cline{2-9}
    \cmidrule(lr){2-3} \cmidrule(lr){4-5} \cmidrule(lr){6-7} \cmidrule(lr){8-9}
    % & acc@0.25(\%) & acc@0.5(\%) & acc@0.25(\%) & acc@0.5(\%) & acc@0.25(\%) & acc@0.5(\%) & acc@0.25(\%) & acc@0.5(\%) \\
    & \makecell[c]{acc@\\0.25(\%)} &  \makecell[c]{acc@\\0.5(\%)}  & \makecell[c]{acc@\\0.25(\%)} & \makecell[c]{acc@\\0.5(\%)} & \makecell[c]{acc@\\0.25(\%)} &  \makecell[c]{acc@\\0.5(\%)}  & \makecell[c]{acc@\\0.25(\%)} & \makecell[c]{acc@\\0.5(\%)} \\
    \midrule
    CVM-Net & 4.73 & 0.51 & 5.09 & 0.87 & 20.14 & 3.29 & 20.04 & 3.47  \\
    RK-Net & 7.40 & 0.82 & 8.67 & 0.98 & 19.22 & 2.67 & 19.94 & 3.03  \\
    L2LTR & 10.69 & 2.16 & 12.24 & 1.84 & 38.95 & 6.27 & 38.68 & 5.96   \\
    Polar-SAFA & 20.66 & 3.19 & 19.18 & 2.71 & 37.41 & 6.58 & 36.19 & 6.39    \\
    TransGeo & 21.17 & 2.88 & 21.67 & 3.25 & 35.05 & 6.37 & 34.78 & 5.42   \\
    SAFA & 22.20 & 3.08 & 20.59 & 3.25 & 37.41 & 6.58 & 36.19 & 6.39   \\
    Sample4Geo   & 5.75 & 1.21 & 6.18 &  0.56 & 6.75 & 1.61 & 7.04 & 1.08 \\
    ConGeo & 30.27 & 5.52 & 29.59 &  5.09 & 34.94 & 6.66 & 30.60 & 5.60  \\
    DetGeo  & 45.43 &42.24 & 46.70 & \underline{43.99} & \underline{61.97} & \underline{57.66} & \underline{59.81} & \underline{55.15}  \\
    VAGeo  & \underline{47.38} & \underline{43.68} & \underline{47.13} & 43.88 & 58.79 & 54.98 & 56.66 & 52.55 \\
    % \hline
    \midrule
    AFGeo(Ours) & \textbf{52.31} & \textbf{49.64} & \textbf{52.87} & \textbf{48.65} & \textbf{66.29} & \textbf{63.10} & \textbf{64.68} & \textbf{59.70} \\
    \bottomrule
\end{tabular}
\caption{Evaluation on CVOGL. Suboptimal results are underlined and best results are shown in bold.}
\label{tab:CVOGL}
\end{table*}
% CVOAM的贡献很大，anchor-base应该也能用 位置编码看作附带的

\begin{table}[t]
\centering
\begin{tabular}{l|cc|cc}
    \toprule
    \multirow{4}{*}{ Method } & \multicolumn{4}{c}{Ground→Drone} \\
    % \cline{2-9}
    \cmidrule(lr){2-5} 
    & \multicolumn{2}{c}{ Test } & \multicolumn{2}{c}{ Validation } \\
    % \cline{2-9}
    \cmidrule(lr){2-3} \cmidrule(lr){4-5} 
    % & acc@0.25(\%) & acc@0.5(\%) & acc@0.25(\%) & acc@0.5(\%) & acc@0.25(\%) & acc@0.5(\%) & acc@0.25(\%) & acc@0.5(\%) \\
    & \makecell[c]{acc@\\0.25(\%)} &  \makecell[c]{acc@\\0.5(\%)}  & \makecell[c]{acc@\\0.25(\%)} & \makecell[c]{acc@\\0.5(\%)} \\
    \midrule
    RK-Net & 13.84 & 1.47 & 12.85 & 1.58  \\
    L2LTR & 19.35 & 3.81 & 18.57 & 3.61   \\
    TransGeo & 27.02 & 4.48 & 28.12 & 7.23 \\
    Sample4Geo  & 48.70 & 7.49 & 54.91 &  8.26 \\
    ConGeo & 63.83 & 11.44 & 68.29 &  9.72  \\
    DetGeo & 74.59 & 63.51 & 68.29 & 59.95 \\
    VAGeo & \underline{76.49} & \underline{72.97} & \underline{74.31} & \underline{68.75} \\
    % \hline
    \midrule
    AFGeo(Ours) & \textbf{78.65} & \textbf{74.86} & \textbf{74.54} & \textbf{69.68}  \\
    \bottomrule
\end{tabular}
\caption{Evaluation on G2D. Suboptimal results are underlined and best results are shown in bold.}
\label{tab:G2D}
\end{table}

\section{Experiment}
\label{sec:experiment}

\subsection{Dataset and Evaluation Metrics}
\textbf{CVOGL Dataset.} 
% We evaluate our method on the standard CVOGL~\cite{DBLP:journals/tgrs/SunYKFFLLZP23}. 
CVOGL consists of two types of cross-view object geo-localization tasks: Task 1 (Ground → Satellite), where ground-view images are used as queries and satellite images as references; and Task 2 (Drone → Satellite), where drone images are used as queries and satellite images as references. The object of interest in a query image is indicated by a clicked point, while the corresponding object in the reference image is annotated with a bounding box. CVOGL contains 6,239 query-reference pairs for each task, with 4,343 for training, 923 for validation, and 973 for testing.

\textbf{G2D Dataset.} 
G2D is constructed for Ground-to-Drone cross-view object geo-localization by leveraging images from the CVOGL dataset. Ground images are used as queries and drone images as references. G2D contains 2753 query-reference pairs for each task, with 1951 for training, 432 for validation, and 370 for testing.

% \textbf{Evaluation metrics.} We adopt acc@t as the evaluation metric, where t is Intersection over Union (IoU) threshold between the predicted and ground-truth boxes, measuring the accuracy of object localization. We consider the prediction to be
% correct when the IoU is greater than or equal to the threshold.

\subsection{Implementation Details}
We employ pre-trained ResNet-50 as the shared backbone for the two branches. The model is optimized using SGD with an initial learning rate of 0.01 and trained for 40 epochs on NVIDIA A100 GPUs. 
% We are surprised to find that our backbone network is remarkably simple, yet still achieves state-of-the-art performance, demonstrating the effectiveness of our method.

\subsection{Comparative Experiment}
We compare our method with the current state-of-the-art methods on CVOGL and G2D. The experimental results are shown in Table~\ref{tab:CVOGL} and Table~\ref{tab:G2D}. Note that the authors of VAGeo~\cite{10888758} have not released the source code and the results reported in the table are reproduced by us based on the descriptions in the original paper. According to Table~\ref{tab:CVOGL} and Table~\ref{tab:G2D}, our method achieves significant gains on both Ground→Satellite, Drone→Satellite and Ground→Drone object localization tasks. Notably, under the more challenging Ground→Satellite setting with severe viewpoint changes, our approach improves acc@0.5 by 13.64\% on the test set and 10.5\% on the validation set. These results demonstrate the superiority of our anchor-free design in various cross-view object geo-localization scenarios.

\subsection{Ablation Study}
We conduct an ablation study to assess the contributions of GPE and CVOAM, as shown in Table~\ref{tab:ablation_CVOGL} and Table~\ref{tab:ablation_G2D}. Removing both yields the worst performance, while adding either module individually improves results. The best performance occurs when both are combined, highlighting the importance of jointly using GPE and CVOAM for robust object geo-localization.

\begin{table*}[t]
\centering
\begin{tabular}{cc|cc|cc|cc|cc}
    \toprule
    \multirow{4}{*}{ GPE } & \multirow{4}{*}{ CVOAM } & \multicolumn{4}{c}{Ground→Satellite} & \multicolumn{4}{c}{Drone→Satellite} \\
    \cmidrule(lr){3-6} \cmidrule(lr){7-10}
    & & \multicolumn{2}{c}{ Test } & \multicolumn{2}{c}{ Validation } &  \multicolumn{2}{c}{ Test } & \multicolumn{2}{c}{ Validation } \\
    % \cline{2-9}
    \cmidrule(lr){3-4} \cmidrule(lr){5-6} \cmidrule(lr){7-8} \cmidrule(lr){9-10}
    % & acc@0.25(\%) & acc@0.5(\%) & acc@0.25(\%) & acc@0.5(\%) & acc@0.25(\%) & acc@0.5(\%) & acc@0.25(\%) & acc@0.5(\%) \\
    & & \makecell[c]{acc@\\0.25(\%)} &  \makecell[c]{acc@\\0.5(\%)}  & \makecell[c]{acc@\\0.25(\%)} & \makecell[c]{acc@\\0.5(\%)} & \makecell[c]{acc@\\0.25(\%)} &  \makecell[c]{acc@\\0.5(\%)}  & \makecell[c]{acc@\\0.25(\%)} & \makecell[c]{acc@\\0.5(\%)} \\
    \midrule
    \xmark & \xmark & 49.64 & 47.28 & 48.00 & 44.85 & 51.59 & 49.33 & 47.35 & 43.88 \\
    \cmark & \xmark & 51.90 & 49.33 & 48.75 & 45.40 & 52.93 & 51.18 & 50.16 & 46.80 \\
    \xmark & \cmark & 50.98 & 48.10 & 52.33 & 48.00 & 65.70 & 62.28 & 61.65 & 57.85 \\
    \cmark & \cmark & \textbf{52.31} & \textbf{49.64} & \textbf{52.87} & \textbf{48.65} & \textbf{66.29} & \textbf{63.10} & \textbf{64.68} & \textbf{59.70} \\
    \bottomrule
\end{tabular}
\caption{Ablation for GPE and CVOAM on CVOGL. Best results are shown in bold.}
\label{tab:ablation_CVOGL}
% CVOAM的贡献很大，anchor-base应该也能用 位置编码看作附带的

\end{table*}

\begin{table}[t]
\centering
\begin{tabular}{cc|cc|cc}
    \toprule
    \multirow{4}{*}{ GPE } & \multirow{4}{*}{ CVOAM } & \multicolumn{4}{c}{Ground→Drone} \\
    \cmidrule(lr){3-6} 
    & & \multicolumn{2}{c}{ Test } & \multicolumn{2}{c}{ Validation } \\
    % \cline{2-9}
    \cmidrule(lr){3-4} \cmidrule(lr){5-6} 
    % & acc@0.25(\%) & acc@0.5(\%) & acc@0.25(\%) & acc@0.5(\%) & acc@0.25(\%) & acc@0.5(\%) & acc@0.25(\%) & acc@0.5(\%) \\
    & & \makecell[c]{acc@\\0.25(\%)} &  \makecell[c]{acc@\\0.5(\%)}  & \makecell[c]{acc@\\0.25(\%)} & \makecell[c]{acc@\\0.5(\%)} \\
    \midrule
    \xmark & \xmark & 74.01 & 69.68 & 70.81 &  65.15 \\
    \cmark & \xmark & 76.49 & 72.97 & 72.84 & 68.15  \\
    \xmark & \cmark & 77.84 & 73.32 & 73.48 & 69.44 \\
    \cmark & \cmark & \textbf{78.65} & \textbf{74.86} & \textbf{74.54} & \textbf{69.68} \\
    \bottomrule
\end{tabular}
\caption{Ablation for GPE and CVOAM on G2D. Best results are shown in bold.}
\label{tab:ablation_G2D}
\end{table}

\subsection{Visual Analysis}
\begin{figure}[t]
\centering
\includegraphics[width=\linewidth]{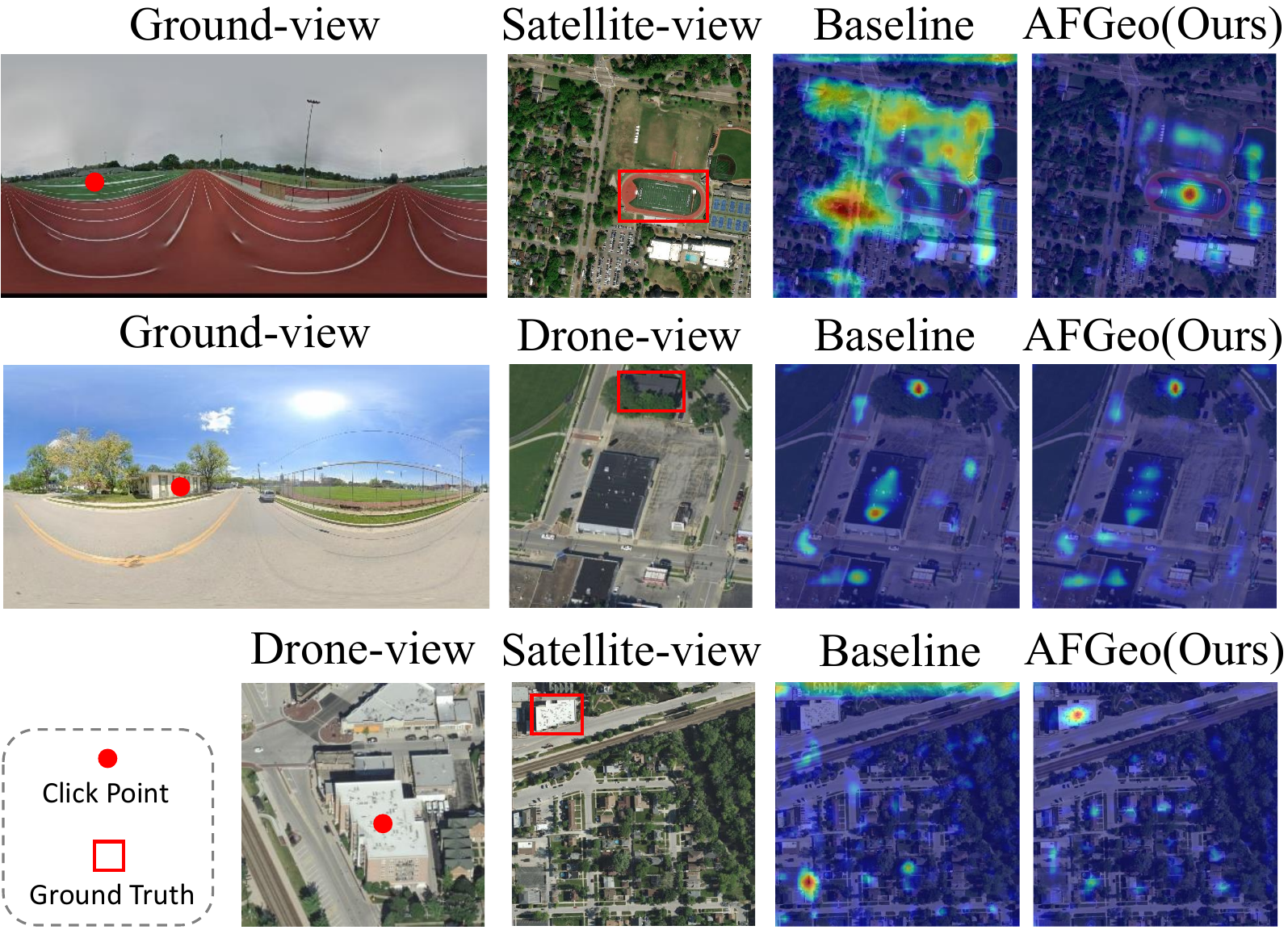}
\caption{Heatmap visualization of Baseline (our framework without GPE and CVPAM) and the proposed AFGeo. The red dot denotes the click point and the red box denotes the ground-truth bounding box.}
\label{fig:heatmap}
\end{figure}

\begin{figure*}[t]
\centering
\includegraphics[width=.95\linewidth]{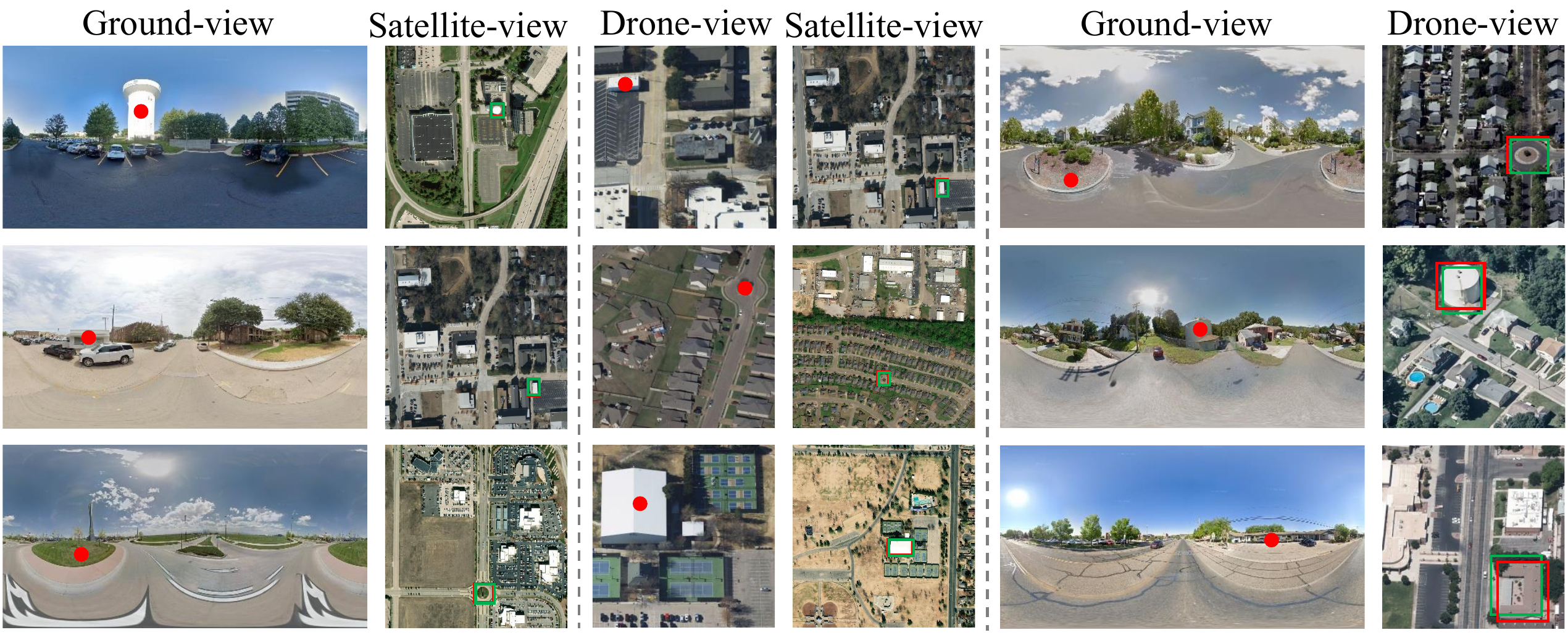}
\caption{Object localization results by our AFGeo. The red and green bounding boxes denote the ground-truth and predicted results, respectively.}
\label{fig:bbox}
\end{figure*}

The heatmaps in Fig.~\ref{fig:heatmap} illustrate the effectiveness of our method across different scenarios. In the Ground→Satellite scenario, the Baseline (our framework without GPE and CVPAM) incorrectly attends to irrelevant regions such as roads and forests, whereas AFGeo accurately highlights the query object—the football field—along with its adjacent contextual structures, including semantically related areas such as tennis and baseball courts. In the Ground→Drone scenario, although the Baseline captures the queried building, it also overemphasizes several visually similar buildings, while our method mitigates such distractions and maintains focus on the true target. In the Drone→Satellite scenario, the Baseline fails to locate the correct building, whereas our method successfully identifies the target. These observations demonstrate that the proposed GPE and CVPAM not only guide the model to accurately attend to the query object but also exploit its surrounding semantically consistent context across views, thereby improving both the robustness and accuracy of object localization. Furthermore, Fig.~\ref{fig:bbox} presents qualitative results under different cross-view scenarios, where the high overlap between the predicted and ground-truth bounding boxes highlights the precise localization performance achieved by our AFGeo.

\section{Conclusion}
We first introduce an anchor-free architecture in cross-view object geo-localization (AFGeo), incorporating GPE to provide a more robust positional prior and CVOAM to associate consistent object semantics across views, thereby jointly enhancing localization performance. AFGeo is a lightweight and efficient model structure, and we hope this work can serve as a reference for deploying cross-view object geo-localization systems in resource-constrained environments.

\bibliography{aaai2026}


\end{document}